\documentclass{llncs}
\usepackage{graphicx}

\usepackage{subfigure}
\usepackage{array}
\usepackage{multirow}

\usepackage{tikz}
\usepackage{comment}
\usepackage{amsmath,amssymb} 
\usepackage{color}

\usepackage[accsupp]{axessibility}

\begin{document}
\pagestyle{headings}
\mainmatter
\def\ECCVSubNumber{6837}  

\title{SGBANet: Semantic GAN and  Balanced Attention Network for Arbitrarily Oriented Scene Text Recognition} 


%
\author{Dajian Zhong  \inst{1}\and
Shujing Lyu \inst{1}\thanks{Corresponding author} \and
Palaiahnakote Shivakumara\inst{2} \and
Bing Yin \inst{3} \and
Jiajia Wu \inst{3} \and
Umapada Pal \inst{4} \and
Yue Lu \inst{1}}


\institute{
Shanghai Key Laboratory of Multidimensional Information Processing, East China Normal University, Shanghai, China \\
\email{djzhong@stu.ecnu.edu.cn,\{sjlv, ylu\}@cs.ecnu.edu.cn}\\
\and
Faculty of Computer Science and Information Technology, University of Malaya, Kuala Lumpur, Malaysia \\
\email{shiva@um.edu.my}\\
\and
iFLYTEK Research, iFLYTEK, Hefei, China \\
\email{\{bingyin, jjwu\}@iflytek.com}\\
\and
CVPR Unit, Indian Statistical Institute, Kolkata, India \\
\email{umapada@isical.ac.in}}

%


\maketitle

\begin{abstract}
Scene text recognition is a challenging task due to the complex backgrounds and diverse variations of text instances. In this paper, we propose a novel Semantic GAN and Balanced Attention Network (SGBANet) to recognize the texts in scene images. The proposed method first generates the simple semantic feature using Semantic GAN and then recognizes the scene text with the Balanced Attention Module. The Semantic GAN aims to align the semantic feature distribution between the support domain and target domain. Different from the conventional image-to-image translation methods that perform at the image level, the Semantic GAN performs the generation and discrimination on the semantic level with the Semantic Generator Module (SGM) and Semantic Discriminator Module (SDM). For target images (scene text images), the Semantic Generator Module generates simple semantic features that share the same feature distribution with support images (clear text images). The Semantic Discriminator Module is used to distinguish the semantic features between the support domain and target domain. In addition, a Balanced Attention Module is designed to alleviate the problem of attention drift. The Balanced Attention Module first learns a balancing parameter based on the visual glimpse vector and semantic glimpse vector, and then performs the balancing operation for obtaining a balanced glimpse vector. Experiments on six benchmarks, including regular datasets, i.e., IIIT5K, SVT, ICDAR2013, and irregular datasets, i.e., ICDAR2015, SVTP, CUTE80, validate the effectiveness of our proposed method.

\end{abstract}

\section{Introduction}

Scene text recognition has attracted growing research attention as a great need in real-life applications such as visual question answering \cite{BitenTMBRJVK19}, license plate recognition \cite{YueKLSZ20}, driver-less vehicles \cite{LeBCH20}. Although previous works \cite{aberdam2021sequence,baek2019wrong,kang2020unsupervised,liao2020mask,zhang2020adaptive,bhunia2021metahtr} have achieved great success, scene text recognition is still a challenging task because scene text images can be affected by multiple factors, such as complex background, arbitrarily shaped text, non-uniform spacing, etc.

\begin{figure}[tp]\centering
\begin{tabular}{m{2.2cm}<{\centering}m{1.8cm}<{\centering} m{1.8cm}<{\centering} m{1.8cm}<{\centering} m{1.8cm}<{\centering}}
\hline
\noalign{\smallskip}
Input &    \includegraphics[width=1.8cm,height=1cm]{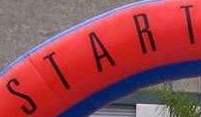}   &     \includegraphics[width=1.8cm,height=1cm]{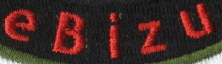}   &       \includegraphics[width=1.8cm,height=1cm]{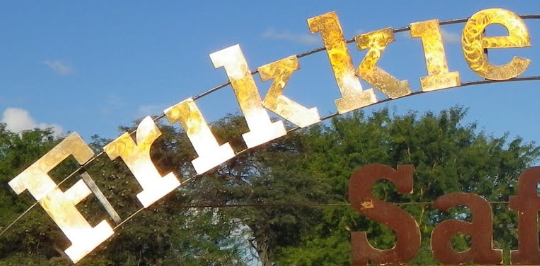}  &  \includegraphics[width=1.8cm,height=1cm]{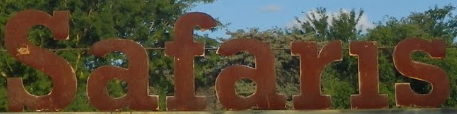}   \\ 

\hline

GT   &    START      &   EBIZU  &  FRIKKIE     & SAFARIS\\ 
\noalign{\smallskip}
Baek et al. \cite{baek2021if} &    STAR\textcolor{red}{I} &       \textcolor{red}{B}BIZU     &   F\textcolor{red}{A}R\textcolor{red}{?}KI\textcolor{red}{NG}  &      SA\textcolor{red}{B}A\textcolor{red}{tt}S \\
 
\noalign{\smallskip}
Fang et al. \cite{fang2021read}   &   START    &  \textcolor{red}{B}B\textcolor{red}{Z}ZU    &  FRIKK\textcolor{red}{L}E  &  SAFA\textcolor{red}{?N}S    \\ 

\hline
\end{tabular}
\caption{Challenges of scene text recognition. GT represents the GroundTruth. Miss-recognized characters are marked in red and `?' represents the missed character.}
\label{fig:vis_challenge}
\end{figure}

With the development of Convolutional Neural Network (CNN) \cite{JaderbergSVZ14b,jaderberg2016reading} and attention-based recognizer \cite{cheng2018aon,li2019show}, text instances with simple background can be well recognized. Although, several models are proposed for addressing challenges of scene text recognition especially for regular text images, achieving good performance for arbitrarily oriented and irregularly shaped scene images is still considered as an open challenge. The key reason is that every character in the above two cases has a different orientation and shape in contrast to regular text. It can be observed from Fig. \ref{fig:vis_challenge} that the existing methods \cite{baek2021if,fang2021read} do not recognize the characters correctly for arbitrarily shaped text and text with complex backgrounds. Therefore, designing a robust method for recognizing arbitrarily shaped text is still a challenging task that remains to be solved.

Since it is easy to recognize clear images while those complex images are hard to recognize, there comes a question is it possible to transform complex images into simple ones? We can get inspiration from the existing works. Wang et al. \cite{wang2020scene} proposed a scene text dataset that contains paired real low-resolution and high-resolution images in the wild. Then the TSRNet is developed to reconstruct the high-resolution images for scene text images and recognize them. In this situation, the complex images (low-resolution images) can be transformed into clear images (high-resolution images). However, paired scene text images are needed in this method and it is hard to obtain these images and annotations. Besides, since the method can only deal with low-resolution images, scene text images with complex backgrounds are hard to recognize. Luo et al. \cite{luo2021separating} introduced the generative adversarial network \cite{goodfellow2014generative,yang2019controllable} to remove the backgrounds while retaining the text content and then recognize the new reconstructed images. It satisfies the need for more scenarios, but it combines two networks into one. One network is for generating new images and another one is to recognize them. The method is time-consuming. The above mentioned methods used GAN for scene text recognition, but none of them used GAN as semantic generator and discriminator for scene text recognition. It motivated us to propose a novel model that explores GAN for extracting semantic features.

In this work, we introduce the Semantic GAN which consists of a Semantic Generator Module and a Semantic Discriminator Module. The Semantic Generator Module directly generates the semantic features upon the convolutional features for complex scene text images. Then the Semantic Discriminator Module distinguishes the semantic features between clear images and complex scene text images. In this way, the semantic feature distribution can be aligned and the generated features share more characteristics with those clear images and thus can be easier to recognize. In this framework, paired images and an extra image-to-image translation network are not needed. 

Besides, the attention drift problem is first raised in \cite{cheng2017focusing}. They proposed the FANet method to draw back the drifted attention with a focusing attention mechanism. It first computes the attention center of each predicted label and then generates the probability distributions on the attention regions. Though the FANet method works well, the complex operations cost huge computations. In our work, we propose a Balanced Attention Module, which directly utilizes the convolutional features to correct the attention weights on the semantic features. The Balanced Attention Module significantly alleviates the problem of attention drift.

The Semantic GAN is capable of aligning the distribution between the support domain and target domain and the Balanced Attention is designed to address the problem of attention drift, thus alleviating the challenges. Our main contributions can be summarized as follows:

\begin{enumerate}

    \item We propose a novel text recognition network, called SGBANet, which consists of Semantic GAN and Balanced Attention Module. The proposed method integrates GAN into the recognition network without introducing an extra image-to-image translation network, thus reducing the time complexity.

    \item We introduce the Semantic GAN, which consists of a Semantic Generator Module and a Semantic Discriminator Module. To the best of our knowledge, it is the first time to use the Semantic Generator and Discriminator to generate semantic features for overcoming the challenge of scene text recognition.
    
    \item We design a Balanced Attention Module, which automatically learns a balancing parameter based on the convolutional and semantic features. It corrects the attention weights on the semantic features and draws back the drifted attention to some extent.
    
\end{enumerate}  


\section{Related Work} \label{sec:related}

\subsection{Scene Text Recognition}
Scene text recognition aims to decode a character sequence from the scene images. The methods can be categorized into language-free methods and language-based methods. The language-free methods usually utilize convolutional features without consideration of the character dependency, such as segmentation-based methods and CTC-based methods. Segmentation-based methods \cite{lyu2018mask,xing2019convolutional,wan2020textscanner} first segment the character regions and then recognize each character region to form the final character sequence. CTC-based methods \cite{gao2019reading,hu2020gtc,he2016reading} first extract visual features through CNN and then train with RNN and CTC loss to find the most possible combination. However, these methods lack linguistic information and thus easily miss one or two characters. 

The language-based methods mainly employ the attention mechanism \cite{qiao2020seed}. The encoder-decoder architecture makes use of linguistic information and character dependency. To boost the performance, some methods focus on learning a new feature representation. For example, \cite{aberdam2021sequence} proposed a contrastive learning algorithm that first divides each feature map into a sequence of individual elements and performs the contrastive loss \cite{he2020momentum}. Then the learned representation features are fed to the recognizer. Yan et al. \cite{yan2021primitive} proposed a primitive representation learning method that aims to exploit intrinsic representations of scene text images. Others may focus on integrating the rectification module \cite{LuoJS19,shi2018aster,zhan2019esir,yang2019symmetry} to reconstruct normal images for those irregular images. Then the reconstructed images are fed to the encoder-decoder module for further recognition. However, scene text usually has a variety of shapes and sizes. It is difficult for the rectification module to transform all the irregular text instances into regular ones. Since current attention-based methods suffer from the attention drift problem \cite{cheng2017focusing,liao2019scene}, directly decoding upon the convolutional features or linguistic features will degrade the recognition performance. Inspired by \cite{wang2020faclstm} which generates the character center masks to help focus attention on the right position. In this work, we introduce the Balanced Attention Module, which learns a balancing parameter based on the convolutional and semantic features and draws back the drifted attention. Through the method, the attention drift problem can be alleviated to some extent.

\subsection{Generative Adversarial Network}

With the development of GANs \cite{mao2017least,zhu2017unpaired,odena2017conditional,ZhouGXW020}, image-to-image translation has achieved great success. Several methods integrate GANs to generate clear text images for scene text images \cite{fang2019learning}. This inspires the researchers to combine the GAN and recognition network. Thus, recent recognition methods focus on integrating the adversarial learning concept into the recognition network. For example, \cite{zhang2019sequence} introduced a gated attention similarity (GAS) unit to adaptively focus on aligning the distribution of the source and target sequence data. Luo et al. \cite{luo2021separating} introduced the generative adversarial network to generate a simple image without the complex background for each scene text image. However, it costs much computation if we first generate the required text image and then feed them to a recognition network. Since the two networks both contain huge convolution operations that can result in huge computation, our goal is to design a novel Semantic GAN to align the semantic feature distribution where the Semantic Generator Module directly generates the high-level semantic features and the Semantic Discriminator Module distinguishes between the support domain and target domain.

\section{Methodology}\label{sec:method}
In this section, details of SGBANet are presented. We first describe the overall architecture of the proposed method. Then we dissect the Generator module and the Discriminator module. Finally, the Balanced Attention module is introduced.  

\subsection{Overall Architecture}
\begin{figure}
\centering
\includegraphics[height=6cm, width=11cm]{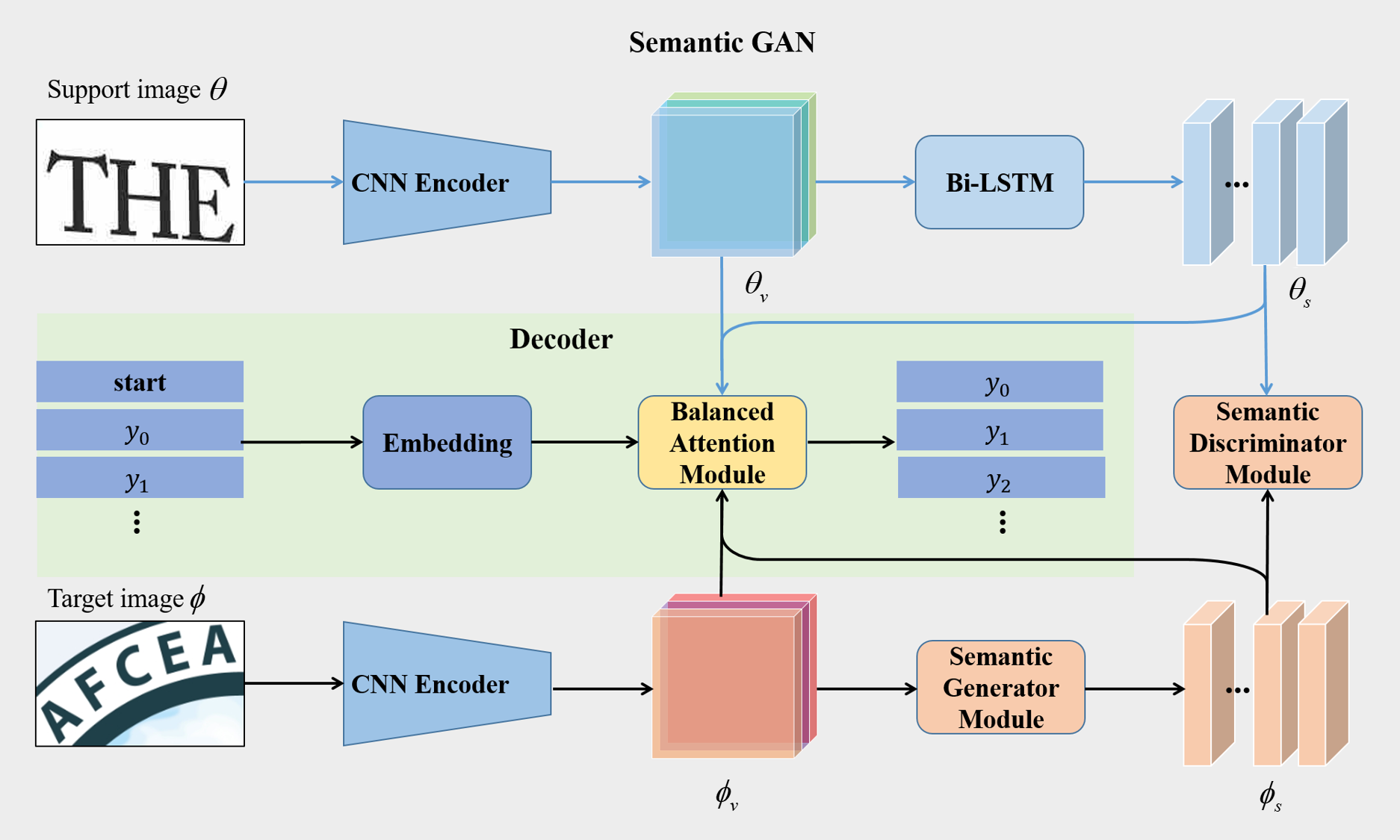}
\caption{Architecture of the proposed SGBANet. The support image $\theta$ and target image $\phi$ are fed to the CNN Encoders, which share the same parameters. The visual convolutional features $\theta_v$ and $\phi_v$ are fed to a Bi-LSTM and the Generator Module, respectively. Moreover, the semantic features $\theta_s$ and $\phi_s$ are fed to the Semantic Discriminator and Balanced Attention for discrimination and recognition. The Embedding module is to make a one-hot embedding on the input labels.}
\label{fig:architecture}
\end{figure}

As can be seen in Fig. \ref{fig:architecture}, the overall architecture of the proposed method consists of the CNN Encoder, the Semantic GAN and the Balanced Attention Module. The CNN Encoder is used to extract the basic convolutional features. The Semantic GAN consists of the Semantic Generator Module(SGM) and the Semantic Discriminator Module (SDM). The Semantic Generator Module is applied to directly generate the high-level semantic features that can be easier to recognize and the Semantic Discriminator Module aims to distinguish between the support semantic features and target semantic features. After the adversarial learning, the semantic feature distribution between the source and target images can be aligned. The Balanced Attention Module is designed to draw back the drifted attention by learning a balancing parameter based on the convolutional and semantic features. Then the balancing operation can be performed to get the balanced glimpse vector.

There are two inputs for the whole architecture. One is the support image $\theta$, which represents the simple image containing pure text instance. Another is the target image $\phi$, which contains text instances with complex backgrounds. Firstly, the two images are fed into the CNN Encoder and visual feature map $\theta_v$ and $\phi_v$ are obtained. Secondly, they are fed to a Bi-LSTM layer and the Semantic Generator Module, respectively, and $\theta_s$ and $\phi_s$ are obtained. Then they are fed to the Semantic Discriminator Module. The Semantic Generator Module and the Semantic Discriminator Module contribute to generating simple semantic features. For the Semantic Generator Module, it generates the semantic feature that the discriminator can't distinguish from the support domain and target domain. At the same time, the Semantic Discriminator Module aims to distinguish them correctly. After the training of Semantic GAN, the Semantic Generator Module successfully generates the simple semantic feature for scene text images, which shares the same feature distribution with those of clear images. And $\phi_v$ and $\phi_s$ together with $\theta_v$ and $\theta_s$ are fed to the Balanced Attention Module for the final recognition.

\subsection{CNN Encoder}

Similar to \cite{cai2021cstr}, we take ResNet-based structure as the backbone network to extract basic visual features. We remove the CBAM module \cite{woo2018cbam} and assemble an extra down-sampling layer. As a result, the size of $\theta_v$ and $\phi_v$ is restored to $\frac{H}{8} \times \frac{W}{8} \times C$, where $H$, $W$ and $C$ represent the height, width and channel of the input images, respectively. Since the size of text instances varies and there is no extra image-to-image network, FPN \cite{lin2017feature} is assembled to make the CNN Encoder capable of extracting different levels of features. The input support image and target image can be fed to a shared CNN Encoder for saving memory and computation cost.

\subsection{Semantic GAN} \label{sec:generator}

The Semantic GAN consists of a Semantic Generator Module and a Semantic Discriminator Module. The Semantic Generator Module directly generates the semantic features for complex scene text images. A Bi-LSTM module is used to extract the semantic features of clear images. Then the Semantic Discriminator Module distinguishes the semantic features between support and target domain. In this way, the semantic feature distribution between the source and target images can be aligned and thus can be easier to recognize.

\subsubsection{Semantic Generator Module} 
For the support image $\theta$, the visual convolutional feature $\theta_v$ is directly fed to a Bi-LSTM layer and the semantic feature $\theta_s$ is obtained. As for the target image $\phi$, the visual convolutional feature $\phi_v$ is fed to the Semantic Generator Module for generating the semantic feature $\phi_s$ so that the new generated feature shares the same feature distribution with those of support images. Different from the structure of the conventional generator that stacks the convolutional layers, the main components of the Semantic Generator Module are Bi-LSTM layers and Fully connected (FC) layers. The generator consists of two basic units and each unit comprises a Bi-LSTM layer followed by two FC layers. Given the input visual convolutional feature $\phi_v$, the Semantic Generator Module generates a new semantic feature $\phi_s$ without encoding the background information. The size of $\phi_s$ is the same as $\theta_s$, which is $\frac{W}{8} \times 256 $.

\begin{figure}
\centering
\includegraphics[height=5cm]{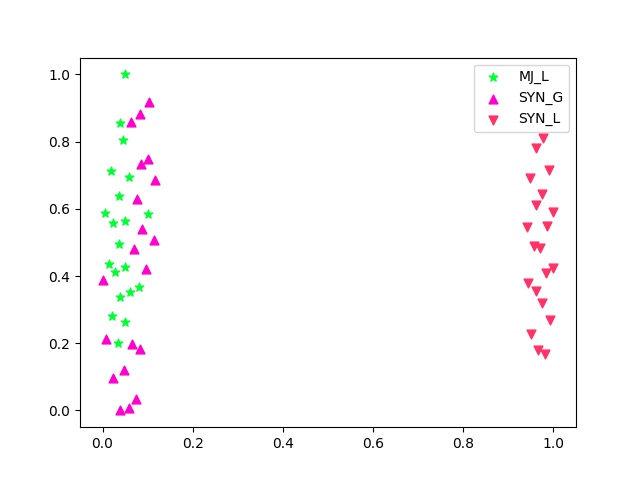}
\caption{T-SNE plot of learned feature representations. `MJ\_L' denotes learned features from Bi-LSTM with sampled MJSynth images. `SYN\_G' and `SYN\_L' denote generated semantic features and learned features from the Semantic Generator Module and Bi-LSTM using sampled SynthText images, respectively.}
\label{fig:gan_effect}
\end{figure}

\subsubsection{Semantic Discriminator Module} \label{sec:discriminator}

The Semantic Discriminator Module is used to discriminate the semantic features $\theta_s$ and $\phi_s$ and the architecture contains no convolutional layers. The Discriminator consists of two basic units and an FC layer. Each unit comprises a Bi-LSTM layer and an FC layer. The first two units are used to reduce the size of input features and the last FC layer is to do the final discrimination. Given the semantic features $\theta_s$ and $\phi_s$, the discriminator distinguishes them between the support image and target image. The output size of the discriminator is $1$. To illustrate the effectiveness of the proposed Semantic GAN, we randomly sample 20 MJSynth images and 20 SynthText images, which represent the clear text images and complex scene text images respectively, and visualize the learned features. As can be seen from Fig. \ref{fig:gan_effect}, the Semantic Generator Module successfully generates the semantic features for the SynthText images that share the same domain with those of MJSynth images.

\subsection{Balanced Attention Module} \label{sec:balance}

\begin{figure}
\centering
\includegraphics[height=6cm]{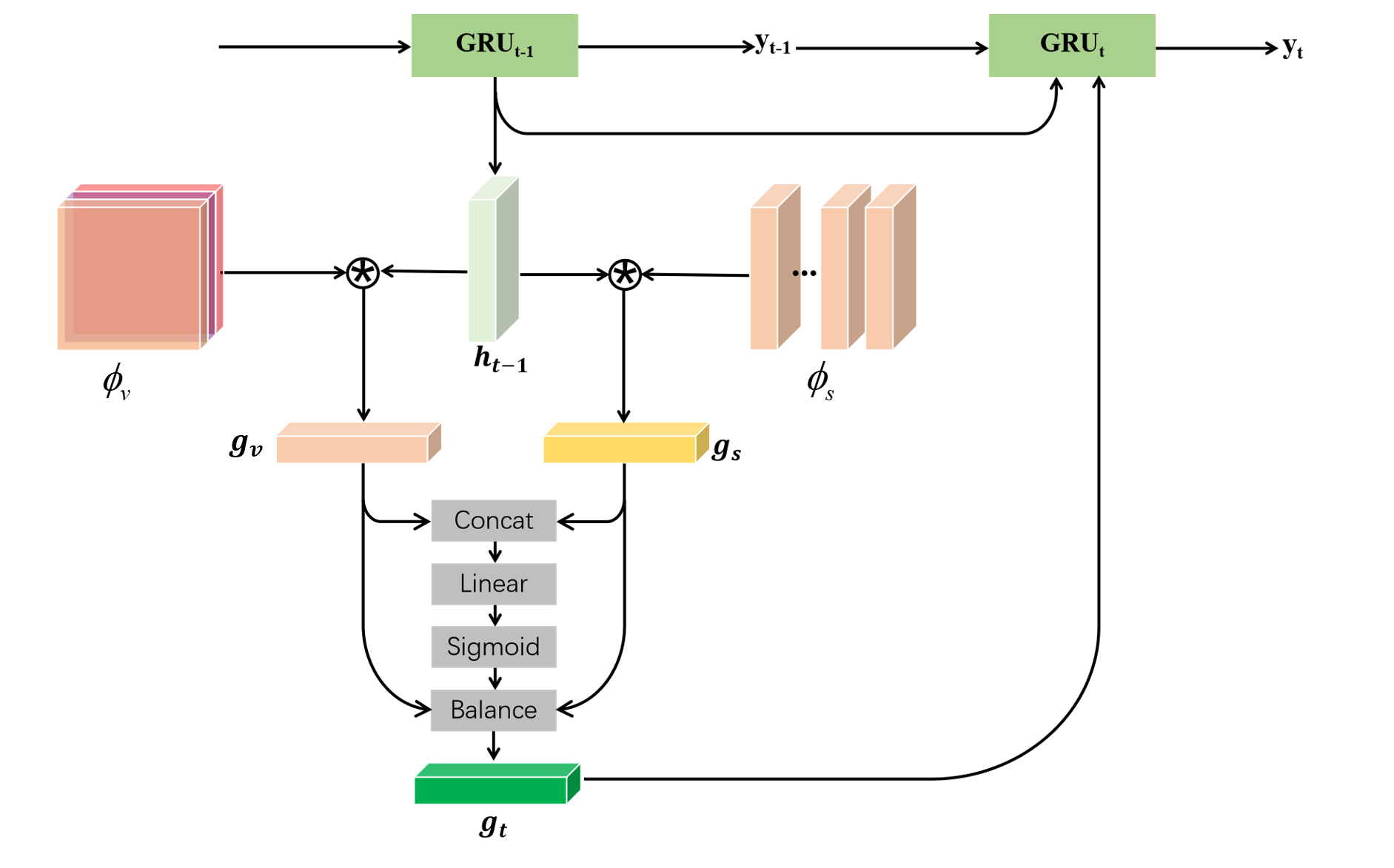}
\caption{Architecture of Balanced Attention Module. The `Balance' operation means making a balance between the two glimpse vectors and it is defined by Equation. (\ref{eq:gt}) and Equation. (\ref{eq:lamda}).}
\label{fig:ba}
\end{figure}

\begin{figure}\centering
\begin{tabular}{m{2.6cm}<{\centering}m{2.6cm}<{\centering} m{2.6cm}<{\centering}}
\includegraphics[width=2.6cm]{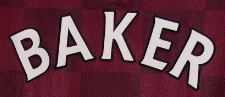} &    
\includegraphics[width=2.6cm]{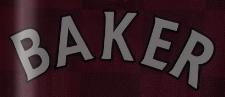} &     
\includegraphics[width=2.6cm]{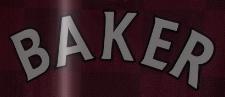} \\
\includegraphics[width=2.6cm]{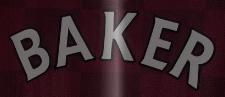} &       
\includegraphics[width=2.6cm,]{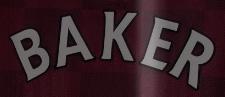} &  
\includegraphics[width=2.6cm]{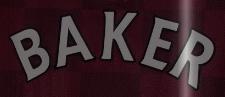} \\ 
\end{tabular}
\caption{An example of the balanced glimpse vector. The first image is the input and the others are attention maps under different time steps.}
\label{fig:ba_vis}
\end{figure}

Since convolutional features contain positions of characters, we can utilize the convolutional features to correct the attention weights on the semantic features. The Balanced Attention Module is designed to draw back the drifted attention and recognize the character sequence. It works iteratively for $T$ steps, producing a target sequence of length $T$, denoted by ($y_1$, $y_2$, ..., $y_T$ ). As can be seen in Fig. \ref{fig:ba}, at time step $t$, the output $y_t$ for target images is defined by:

\begin{equation}\label{eq:yt}
{y_t} = Softmax ({W_{out}}{h_t}+{b_{out}})
\end{equation}

where $W_{out}$ and $b_{out}$ are trainable parameters and $h_t$ is the hidden state at the time step $t$. $h_t$ is updated as follows:

\begin{equation}\label{eq:ht}
{h_t} = GRU({y_{t - 1}},h_{t - 1},{g_t})
\end{equation}

where $y_{t - 1}$ is the output at time step $t-1$ and $g_t$ represents the glimpse vector defined by:

\begin{equation}\label{eq:gt}
g_t  = \lambda g_v  + (1 - \lambda )g_s \begin{array}{*{20}c}
   {}  \\
\end{array},\begin{array}{*{20}c}
   {}  \\
\end{array}\lambda  \in R^n 
\end{equation}

where $\lambda$ is learnable parameters. $g_v$ and $g_s$ are internal glimpse vectors. They are computed as follows:
\begin{equation}\label{eq:lamda}
\lambda  = Sigmoid(W_{\lambda} cat(g_v ,g_s ) + b_{\lambda} )
\end{equation}

\begin{equation}\label{eq:gc}
g_v  = \sum\limits_{i = 1}^n {\alpha _{t,i}^v  \times \phi_{v,i} } 
\end{equation}

\begin{equation}\label{eq:gs}
g_s  = \sum\limits_{i = 1}^n {\alpha _{t,i}^s  \times \phi_{s,i} } 
\end{equation}

where $\alpha _t^v$ and $\alpha _t^s$ are attention weights, which are defined by:

\begin{equation}\label{eq:atc}
\alpha _{t,i}^v  = \frac{{\exp (e_{t,i}^v )}}{{\sum\limits_{j = 1}^n {\exp (e_{t,j}^v )} }}
\end{equation}

\begin{equation}\label{eq:ats}
\alpha _{t,i}^s  = \frac{{\exp (e_{t,i}^s )}}{{\sum\limits_{j = 1}^n {\exp (e_{t,j}^s )} }}
\end{equation}

\begin{equation}\label{eq:etc}
e_{t,i}^v  = W_c  \times \tanh (W_h h_{t - 1}  + W_v* AP(\phi_v)  + b_v )
\end{equation}

\begin{equation}\label{eq:ets}
e_{t,i}^s  = W_c  \times \tanh (W_h h_{t - 1}  + W_s*\phi_s  + b_s )
\end{equation}

where $W_c$, $W_h$, $W_v$, $W_s$, $b_v$ and $b_s$ are trainable parameters and AP represents average pooling on the height of $\phi_v$. Thus the size of $AP(\phi_v)$ is restored into  $\frac{W}{8} \times C $ (C is set to 256), which is the same as that of $\phi_s$. For the recognition of support images, the $\phi_v$ and $\phi_s$ can be replaced by $\theta_v$ and $\theta_s$, and the above operations can be performed in the same way. As can be seen in Fig. \ref{fig:ba_vis}, the balanced glimpse vector learns an accurate attention weight and guide the recognition of characters.

In the inference stage, since the input image can be a clear image or a complex scene text image, we feed the convolutional feature $\phi_v$ to the Bi-LSTM and Semantic Generator Module both. Then the two learned features are fed to the Balanced Attention Module for recognition and two character sequences are produced. Finally, we choose the character sequence with a higher confidence score as the final result and the other one will be discarded.

\subsection{Training}

The objective function $L$ of our proposed SGBANet consists of two parts: the recognition loss $L_R$ and the GAN loss $L_G$, as defined by:

\begin{equation}\label{eq:loss}
L = L_R  + \beta L_G 
\end{equation}

where $\beta$ is the coefficient used to balance the importance of the recognition network and GAN network. We set $\beta$ to 0 during the pre-training stage and 1 during the fine-tuning stage. The GAN loss is defined by:

\begin{equation}\label{eq:G}
L_G  = L_g  + \lambda L_d 
\end{equation}

\begin{equation}\label{eq:g}
L_d  = E[\max (0,1 - D(\theta_s ))] + E[\max (0,1 + D(G(\phi_v ))]
\end{equation}

\begin{equation}\label{eq:d}
L_g  =  - E[D(G(\phi_v )]
\end{equation}

where $G( \cdot )$ and $D( \cdot )$ denote the Semantic Generator Module and the Semantic Discriminator Module. $\theta_s$ and $\phi_v$ are semantic feature and visual convolutional feature of support image and target image, respectively.

As it is hard to train the recognition network combined with GAN, joint training of the recognition network and GAN will easily cause the instability of recognition loss. To make the whole network stable and robust enough, we have tried several loss functions for the GAN network, such as the original GAN loss \cite{goodfellow2014generative}, Wasserstein GAN Loss \cite{arjovsky2017wasserstein}, and Hinge loss \cite{xie2021dg}. Additionally, we have tried several architectures of the Semantic Generator Module and the Semantic Discriminator Module, such as integrating convolution layers, LSTM layers and FC layers. As a result, we find that the Hinge loss, together with the architecture reported in Section \ref{sec:generator} and Section \ref{sec:discriminator}, gets a robust training.

\section{Experiments}\label{sec:exp}
\subsection{Datasets}
We train the proposed method SGBANet with the public available synthetic datasets, $i.e.$ SynthText \cite{gupta2016synthetic} and MJSynth \cite{jaderberg2014synthetic} without fine-tuning on individual scene text datasets and lexicons. We evaluate the performance on the six widely used benchmarks, including three regular text datasets (IIIT5K, ICDAR2013, SVT), and three irregular text datasets (ICDAR2015, SVTP, CUTE80). Details of the datasets are as follows.

\textbf{IIIT5K} \cite{mishra2012top} contains 3000 test images cropped from natural scene images. Most of the text instances are regular with the horizontal layout. 

\textbf{ICDAR2013 (IC13)} \cite{karatzas2013icdar} has 1015 test images. The dataset only contains horizontal text instances.

\textbf{Street View Text (SVT)} \cite{wang2011end} consists of 647 word patches cropped from Google Street View for testing. This dataset contains blur, noise and low-resolution text images. 

\textbf{ICDAR2015 (IC15)} \cite{karatzas2015icdar} contains 2077 test images collected by Google Glasses. Most of the text instances are irregular (oriented, perspective or curved). We discard the vertical text images, which results in 2002 test images. 

\textbf{Street View Text Perspective (SVTP)} \cite{phan2013recognizing} contains 645 cropped images from side view angle snapshots in Google Street View. Most of the images are perspective distorted.

\textbf{CUTE80} \cite{risnumawan2014robust} contains 288 images cropped from high-resolution scene text images. Most of the images contain curved text.

\subsection{Implementation Details}
The proposed method is implemented by using PyTorch. All the experiments are conducted on an NVIDIA Tesla V100 GPU with 32 GB memory. In our experiments, all the input images are rescaled to the size of $64 \times 256$ with the aspect ratio preserved. The character set contains 64 classes, including 10 digits, 52 case-sensitive letters, the $SOS$ token and the $EOS$ token. The maximum sequence length is set to 32. AdaDelta is chosen as the optimizer and the batch size is set to 185. 

The training process of the SGBANet is divided into two stages: the pre-training stage and the fine-tuning stage. In the pre-training stage, as we described in Equation. (\ref{eq:loss}), we set the $\beta$ to 0. Thus, we train the network on the SynthText and MJSynth from scratch for 2 epochs. Since images in MJSynth contain only pure text instances and images in SynthText contain complex backgrounds, images in the MJSynth and SynthText can be considered as support images and target images, respectively. In the fine-tuning stage, $\beta$ is set to 1, and we jointly train the whole network for another 5 epochs.

\subsection{Comparison with State-of-the-art Approaches}
\begin{figure}\centering
\begin{tabular}{m{2.5cm}<{\centering}m{2cm}<{\centering} m{2cm}<{\centering} m{2cm}<{\centering} m{2cm}<{\centering}}
\hline
\noalign{\smallskip}
Input &    \includegraphics[width=2cm,height=1cm]{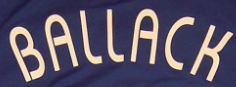}   &     \includegraphics[width=2cm,height=1cm]{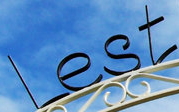}   &       \includegraphics[width=2cm,height=1cm]{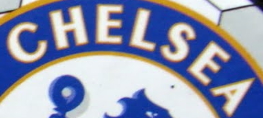}  &  \includegraphics[width=2cm,height=1cm]{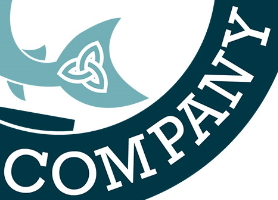}   \\ 
\noalign{\smallskip}
\hline
\noalign{\smallskip}
GT   &    BALLACK    &   LEST  &  CHELSEA     & COMPANY\\ 
\noalign{\smallskip}
 Baek et al. \cite{baek2019wrong}   &    BALLACK   &   LEST    &  CHELSE\textcolor{red}{?}  &  COMP\textcolor{red}{ERS}     \\ 
\noalign{\smallskip}
Baek et al. \cite{baek2021if} &    BALLAC\textcolor{red}{H} &  \textcolor{red}{?} EST  & CHELSEA     & COMPANY\\
\noalign{\smallskip}
SGBANet  &      BALLACK    &   LEST  &  CHELSEA     & COMPANY\\ 
\noalign{\smallskip}

\hline
\end{tabular}
\caption{Qualitative results for text recognition of the existing methods and our proposed method.}
\label{fig:qualitative}
\end{figure}

We evaluate our method on the aforementioned six benchmark datasets and compare it with those state-of-the-art methods. For a fair comparison, all the methods are trained on the SynthText and MJSynth without fine-tuning on individual scene text datasets and no lexicons are used during the inference stage. Table \ref{table:compare_sota} presents the details of the comparison results. It has been shown that our proposed SGBANet achieves the best on four datasets including IIIT5K, ICDAR2013, ICDAR2015 and SVTP, and achieves competitive performance on two datasets including SVT and CUTE80. Note that, for training the Semantic GAN, we have to divide the MJSynth and SynthText into support images and target images, respectively. Only the SynthText dataset is trained on the Semantic Generator Module. As a result, the proposed method doesn't achieve the best on the SVT and CUTE80. If we compare our method with SSDAN\cite{zhang2019sequence}, which exploits the domain adaptation network, we can find that there is a large performance gap between the two methods on the regular text images. Qualitative results for text recognition of the existing methods and our proposed method are presented in Fig. \ref{fig:qualitative}. The existing methods \cite{baek2019wrong,baek2021if} do not recognize the characters correctly, while the proposed method reports correct recognition results. In addition, We have calculated the average FPS of the proposed and existing methods \cite{baek2019wrong,baek2021if} for all the 8 datasets and the results are 6.76, 6.68 and 7.58 for the methods \cite{baek2019wrong}, \cite{baek2021if} and SGBANet, respectively. This shows that our method is faster than the existing methods. The key reason is that the existing methods perform rectification before prediction while the proposed method does not.

\setlength{\tabcolsep}{4pt}
\begin{table}\centering
\begin{center}
\caption{Comparison with state-of-the-art methods.}
\begin{tabular}{lccccccc}
\hline
\multicolumn{1}{c}{\multirow{2}{*}{Methods}}  & \multicolumn{3}{c}{Regular Text} &  & \multicolumn{3}{c}{Irregular Text}  \\
\cline{2-4} \cline{6-8} 
\multicolumn{1}{c}{} & \multicolumn{1}{c}{IIIT5K} & \multicolumn{1}{c}{SVT} & \multicolumn{1}{c}{IC13} & & \multicolumn{1}{c}{IC15}       & \multicolumn{1}{c}{SVTP}   & \multicolumn{1}{c}{CUTE80}     \\ 
\hline
\noalign{\smallskip}
Liao et al. \cite{liao2019scene} &    92.0   &    82.1   &   91.4   &  &   -  &  -    & -  \\
Baek et al. \cite{baek2019wrong} &    87.9   &    87.5   &   92.3   &  &   71.8  &  79.2    & 74.0  \\
Luo et al. \cite{LuoJS19}&    91.2   &    88.3   &   92.4   &  &   68.8  &  76.1    & 77.4  \\
Li et al. \cite{li2019show}&    91.5   &    84.5  &   91.0   &  &   69.2  &  76.4    & 83.3  \\
Shi et al. \cite{shi2016end}&    81.2   &    82.7  &   89.6   &  &   -  &  -    & -  \\
Zhang et al. \cite{zhang2019sequence} &    83.8   &    84.5   &   91.8   &  &   -  &  -    & -    \\
Xie et al. \cite{xie2019aggregation}&    -   &    -   &   -   &  &   68.9  &  70.1    & 82.6    \\
Yue et al. \cite{YueKLSZ20} &    95.3   &    88.1   &   94.8   &  &     77.1    &  79.5 & \textbf{90.3}    \\
Wang et al. \cite{wang2020faclstm}&    90.5   &    82.2   &   -   &  &     -    &  - & 83.3    \\
Long et al. \cite{long2020new} &    93.7   &    88.9   &   92.4   &  &   76.6  &  78.8    & 86.8\\
Wang et al. \cite{wang2020decoupled} &    94.3   &    \textbf{89.2}   &   93.9   &  &   74.5  &  80.0    & 84.4   \\
Luo et al. \cite{luo2020learn} &   -  &   -   &  -   &  &   76.1  &  79.2    & 84.4    \\
Aberdam et al. \cite{aberdam2021sequence} &    82.9   &    87.9   &   -   &  &   -  &  -    & -    \\
Baek et al. \cite{baek2021if} &    92.1   &    88.9   &   93.1   &  &   74.7  &  79.5    & 78.2    \\

\noalign{\smallskip}
\hline
SGBANet &    \textbf{95.4}   &    89.1   &   \textbf{95.1}   &  &   \textbf{78.4}  &  \textbf{83.1}     &  88.2     \\       
\hline
\end{tabular}
\label{table:compare_sota}
\end{center}
\end{table}

\subsection{Ablation Study}
The proposed method mainly consists of three modules: the Semantic Generator Module, the Semantic Discriminator Module and the Balanced Attention Module. To demonstrate the effectiveness of the proposed method, we will first evaluate the performance of individual components and then make a further discussion on the Balance Attention Module. 

\subsubsection{The effectiveness of the individual components.} 

Since the Semantic GAN and Balanced Attention Module are two core components of the proposed method, we combine the baseline method with the individual components and conduct experiments on the six benchmarks. The proposed method without considering Semantic GAN and Balanced Attention Module is considered as the baseline method. The baseline method combined with Semantic GAN is considered as `Baseline+SGAN'. In the same way, the baseline method combined with the two components is considered as `Baseline+SGAN+BA'. It is observed from Table \ref{table:abaltion_module}, `Baseline+SGAN+BA' outperforms the `Baseline' method on all the six benchmarks. It indicates that the Semantic GAN successfully generates simple semantic features that are easier to recognize. In the same way, `Baseline+SGAN+BA' outperforms `Baseline+SGAN', which attests to the effectiveness of the proposed Balance Attention Module.

\setlength{\tabcolsep}{4pt}
\begin{table}\centering
\begin{center}
\caption{Effectiveness of the key components of the proposed method. `SGAN' and `BA' represent the Semantic GAN and Balance Attention.}
\begin{tabular}{lcccccc}
\hline
Datasets &    IIIT5K   &    SVT   &   IC13   &     IC15  &  SVTP     &  CUTE80  \\ 
\hline
\noalign{\smallskip}
Baseline &    90.1   &    84.5   &   91.8     &   70.5  &  76.8     &  80.0    \\

Baseline+SGAN&    92.1   &    86.6   &   91.1     &   72.5  &  77.1     &  80.5    \\

Baseline+SGAN+BA &  \textbf{95.4}   &  \textbf{89.1}  &   \textbf{95.1}   &  \textbf{78.4}  &  \textbf{83.1}  &  \textbf{88.2} \\  
\noalign{\smallskip}
\hline
\end{tabular}
\label{table:abaltion_module}
\end{center}
\end{table}

\subsubsection{Discussions about the Balance Attention Module.} 

The Balance Attention is the core component of our proposed method. The key step of the Balance Attention is the balancing operation, which learns a dynamic balancing parameter and makes a balance between the two glimpse vectors. The details of our balancing operation are defined in Equation. (\ref{eq:gt}) and Equation. (\ref{eq:lamda}). We discuss the effectiveness of the proposed balancing operation and other balancing operations. In the experiment, we evaluate four balancing operations, including `Single', `Add', `CL', and our `Balance' operation. The `Single' operation only uses the glimpse vector $g_v$. The `Add' operation directly makes an add operation on $g_v$ and $g_s$. The `CL' operation generates a new glimpse vector by first concatenating the two glimpse vectors and then feeding it to an FC layer. `Balance' is our proposed balancing operation. As can be seen in Table \ref{table:abaltion_balance}, the `Single' operation gets the worst performance on the six benchmarks. Text instances with complex backgrounds and arbitrary shapes are challenging for the `Single' operation. `Add' and `CL' operations improve the performance and achieve the best on SVT and CUTE80, respectively. Our `Balance' operation further improves the performance and achieves the best performance on four datasets including IIIT5k, IC13, IC15, and SVTP, and there is only a gap of 0.3 on the other two datasets. Thus, our balancing operation gets the overall best performance.

\setlength{\tabcolsep}{4pt}
\begin{table}\centering
\begin{center}
\caption{Evaluation of the balancing operations.}
\begin{tabular}{lcccccc}
\hline
Datasets &    IIIT5K   &    SVT   &   IC13   &     IC15  &  SVTP     &  CUTE80  \\ 
\hline
\noalign{\smallskip}
Single&    92.1   &    86.6   &   91.9     &   72.5  &  77.1     &  80.5    \\

Add&    94.2   &    \textbf{89.4}   &   94.0     &   77.3  &  81.7     &  87.1    \\

CL&    95.1   &    88.3   &   94.1     &   77.2  &  83.0     &  \textbf{88.5}     \\   

Balance&    \textbf{95.4}   &    89.1   &   \textbf{95.1}     &   \textbf{78.4}  & \textbf{ 83.1}     &  88.2     \\   
\noalign{\smallskip}
\hline
\end{tabular}
\label{table:abaltion_balance}
\end{center}
\end{table}

\section{Conclusions}\label{sec:conclusion}

In this paper, we propose a novel Semantic GAN and Balanced Attention Network (SGBANet) for arbitrarily oriented scene Text recognition. The Semantic GAN is designed to align the semantic feature distribution between the support and target domain. The Semantic Generator Module focuses on generating simple semantic features for the scene text images. The Semantic Discriminator Module aims to distinguish the semantic features between the support domain and target domain. Experiments show that the Semantic GAN successfully generates simple semantic features for complex scene text images. The generated simple semantic features share the same feature distribution with those of clear images. Furthermore, to alleviate the problem of attention drift, the Balanced Attention Module is designed. It utilizes the convolutional features to correct the attention weights on the semantic features. A new balancing operation is performed on the two glimpse vectors and a balanced glimpse vector is learned. Ablation study on the baseline method combined with the proposed modules demonstrates the advantages of the designed modules. Extensive experiments on six benchmarks demonstrate the effectiveness of our proposed method.

\section*{Acknowledgements}
This work is supported by the National Key Research and Development Program of China under Grant No. 2020AAA0107903, the National Natural Science Foundation of China under Grant No. 62176091, and the Shanghai Natural Science Foundation of China under Grant No. 19ZR1415900.



\clearpage
%
%
\bibliographystyle{splncs04}
\bibliography{egbib}
\end{document}